\documentclass{article}

\usepackage{microtype}
\usepackage{graphicx}
\usepackage{subcaption}
\usepackage{booktabs} 
\usepackage{enumitem}

\usepackage{url}
\RequirePackage[format=plain,labelformat=simple,labelsep=period,font=small,compatibility=false]{caption}
\RequirePackage[font=footnotesize,skip=3pt,subrefformat=parens]{subcaption}

\usepackage{hyperref}

\usepackage[preprint]{icml2026}

\usepackage{amsmath}
\usepackage{amssymb}
\usepackage{mathtools}
\usepackage{amsthm}

\usepackage{multirow}
\usepackage{graphicx}

\usepackage{colortbl}
\usepackage{xcolor}

\usepackage[capitalize,noabbrev]{cleveref}

\theoremstyle{plain}

\theoremstyle{definition}

\theoremstyle{remark}

\usepackage[textsize=tiny]{todonotes}

\usepackage{xspace}
\newcommand{\NICKNAME}{\textsc{4RC}}
\newcommand{\nickname}{\NICKNAME\xspace}

\begin{document}

\twocolumn[{
  \icmltitle{\nickname: 4D Reconstruction via Conditional Querying Anytime and Anywhere}



  \icmlsetsymbol{equal}{*}

  \begin{icmlauthorlist}
    \icmlauthor{Yihang Luo}{ntu}
    \icmlauthor{Shangchen Zhou}{ntu}
    \icmlauthor{Yushi Lan}{vgg}
    \icmlauthor{Xingang Pan}{ntu}
    \icmlauthor{Chen Change Loy}{ntu}
  \end{icmlauthorlist}

  \icmlaffiliation{ntu}{S-Lab, Nanyang Technological University}
  \icmlaffiliation{vgg}{VGG, University of Oxford}


  \icmlkeywords{4D Reconstruction, Computer Vision}

  \begin{center}
    \vspace{3mm}
    \includegraphics[width=\linewidth]{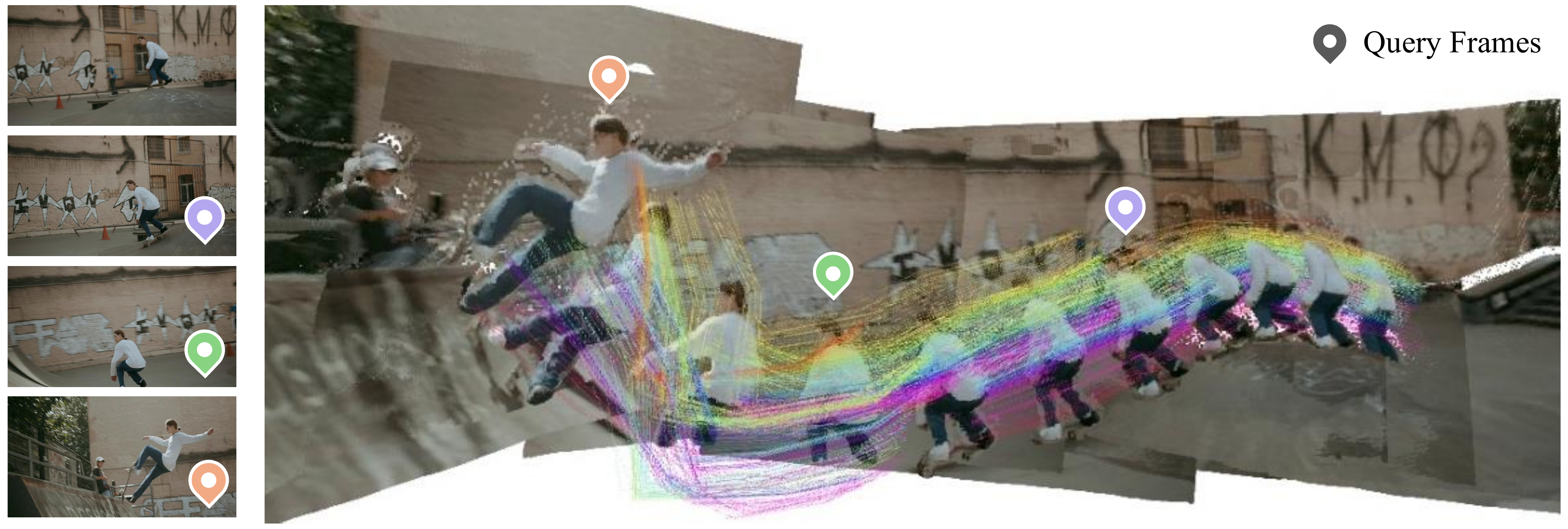}
    \captionof{figure}{\textbf{\nickname}~(pronounced ``ARC'') enables unified and complete \textbf{4}D \textbf{R}econstruction via \textbf{C}onditional querying from monocular videos in a single feed-forward pass. It jointly recovers camera poses and dense per-frame geometry, while supporting flexible querying of dense 3D motion from arbitrary source frames to any target timestamp.}
    \label{fig:teaser}
    \vspace{5mm}
  \end{center}
  }
]


\printAffiliationsAndNotice{}  

\begin{abstract}
We present \nickname, a unified feed-forward framework for 4D reconstruction from monocular videos. Unlike existing methods that typically decouple motion from geometry or produce limited 4D attributes, such as sparse trajectories or two-view scene flow, \nickname learns a holistic 4D representation that jointly captures  dense scene geometry and motion dynamics. 
At its core, \nickname introduces a novel \textit{encode-once, query-anywhere and anytime} paradigm: a transformer backbone encodes the entire video into a compact spatio-temporal latent space, from which a conditional decoder can efficiently query 3D geometry and motion for \textit{any} query frame at \textit{any} target timestamp.
To facilitate learning, we represent per-view 4D attributes in a minimally factorized form, decomposing them into base geometry and time-dependent relative motion.
Extensive experiments demonstrate that \nickname~outperforms prior and concurrent methods across a wide range of 4D reconstruction tasks. \textit{Project Page:} \url{https://yihangluo.com/projects/4RC/}.

\end{abstract}
\section{Introduction}
\label{sec:introduction}

3D reconstruction has seen remarkable progress over the past decades. Classical geometric pipelines such as Structure-from-Motion (SfM)~\cite{schoenberger2016sfm} and Multi-View Stereo (MVS)~\cite{yao2018mvsnet,yao2019recurrent,schoenberger2016mvs} established a solid foundation. 
More recently, learning-based approaches, exemplified by DUSt3R-like pointmap predictor~\cite{wang2024dust3r,mast3r_arxiv24,wang2025cut3r,wang2025vggt,wang2025pi,depthanything3,stream3r2025} have enabled direct feed-forward inference of dense 3D geometry, advancing general-purpose 3D perception in terms of efficiency, scalability, and generalization.
Despite this progress, existing approaches largely focus on static geometry, while real-world scenes are inherently dynamic. A truly general visual perception system must therefore reason not only about 3D structure, but also about how the scene evolves over time.
This motivates the task of \emph{4D reconstruction}, which aims to jointly model 3D geometry and motion.
Such a representation is fundamental for applications ranging from video synthesis~\cite{gu2025das, wu2024cat4d, lee2025editbytrack} and scene understanding to robotics~\cite{lee2025tracegen,huang2026pointworldscaling3dworld}, where reasoning about object trajectories, deformations, and interactions is essential. 
Existing approaches to 4D reconstruction, however, remain fragmented and limited in flexibility. A common strategy decomposes the problem into sequential subtasks, typically separating motion estimation from 3D reconstruction. For example, SpatialTracker~\cite{xiao2024spatialtracker,xiao2025spatialtracker} performs reconstruction and tracking in a staged manner, relying on iterative refinement, and producing only sparse 3D trajectories. MonST3R~\cite{zhang2024monst3r} further requires post-hoc optimization to establish correspondences across time. Although recent feed-forward methods such as ST4RTrack~\cite{st4rtrack2025} and Dynamic Point Map~\cite{sucar2025dynamic} pioneer direct 4D prediction, they are restricted to pairwise views and thus struggle to model long-term and complex motion. Concurrently, TraceAnything~\cite{liu2025traceanythingrepresentingvideo} represents motion using Bézier curves, enabling long-range 3D trajectory tracking, but often at a cost of reduced geometry quality. Any4D~\cite{karhade2025any4d} supports feed-forward 3D reconstruction, but only predicts scene flow for the first frame and is unable to model 3D motion for the remaining frames. V-DPM~\cite{sucar2025vdpm} extends VGGT to 4D, but suffers from slow inference and limited flexibility at inference. 
Motivated by these limitations, we investigate whether a unified, feed-forward model can enable complete and flexible 4D prediction. In this work, we propose \nickname, a unified feed-forward approach for 4D reconstruction from monocular videos.
Unlike previous approaches that require multiple stages, \nickname~learns a holistic and compact 4D representation that jointly encodes scene geometry and motion across the entire video sequence.
This representation serves as a centralized 4D latent from which geometry and motion can be efficiently queried and decoded.
%
Instead of directly reconstructing a full 3D point cloud for each frame at each timestamp, we adopt a compact factorized output formulation. 
Specifically, we represent each frame with a viewpoint-invariant \textit{base geometry} together with time-dependent \textit{relative motion}, parameterized as 3D displacements.
By querying the model at different timestamps, \nickname can recover both geometry and motion information, such as point trajectories between any frame and any target time. This design enables both flexible and efficient 4D reconstruction.

%
Our contributions can be summarized as follows:
%
\begin{itemize}
    \item A unified feed-forward transformer framework for 4D reconstruction from monocular videos, which jointly models 3D geometry and motion within a single network, eliminating the need for auxiliary estimators or per-scene optimization.
    \item An \emph{encode-once, query-anywhere and anytime} paradigm built upon a compact 4D latent representation. This allows our conditional decoder to flexibly retrieve dense 3D geometry and motion for arbitrary query frames at any target timestamp.
    \item A minimally factorized 4D representation that decomposes each frame into a viewpoint-invariant base geometry and time-dependent relative motion, enabling unified and flexible reconstruction of dynamic scenes. 
\end{itemize}

Extensive experiments demonstrate that \nickname{} achieves competitive performance on standard benchmarks across a wide range of 3D and 4D reconstruction tasks, including camera pose estimation, video depth prediction, point cloud reconstruction, 3D point tracking, and dense motion modeling.
\section{Related Work}
\label{sec:related_work}

\noindent \textbf{Feed-forward 3D Reconstruction.} Reconstructing 3D geometry from 2D images is a long-standing problem in computer vision. Traditional pipelines such as SfM~\cite{schoenberger2016sfm} and MVS~\cite{schoenberger2016mvs, yao2018mvsnet,yao2019recurrent} recover camera parameters and dense geometry through multi-stage optimization, achieving strong performance but at high computational cost.
Recent work has shifted toward feed-forward 3D reconstruction, aiming to replace these complex pipelines with a single neural network that directly predicts 3D attributes. DUSt3R~\cite{wang2024dust3r} demonstrates that dense stereo reconstruction can be achieved in one forward pass, while VGGT~\cite{wang2025vggt} further unifies camera pose estimation and depth prediction across multiple views using a transformer backbone. These methods highlight that, given sufficient data and model capacity, feed-forward architectures can effectively solve static 3D reconstruction. Extensions to dynamic settings, such as MonST3R~\cite{zhang2024monst3r}, Pi3~\cite{wang2025pi}, DA3~\cite{depthanything3} and related approaches~\cite{wang2025cut3r,stream3r2025}, jointly estimate camera parameters and per-frame geometry from dynamic data. Despite operating on dynamic scenes, these methods only reconstruct geometry for each view and thus require separate pipelines to explicitly model 3D motion or temporal correspondence. 

\noindent \textbf{Point Tracking.} Modeling motion over time has traditionally been studied through optical flow~\cite{sun2010secrets} and point tracking~\cite{pips}. Optical flow methods~\cite{sun2018pwc,hui18liteflownet,teed2020raft} estimate dense pixel-wise displacements between adjacent frames.
These methods are typically limited to short temporal windows and often suffer from drift errors when applied to long video sequences~\cite{zhou2023propainter}. 
To address long-range correspondence, 2D point tracking methods aim to track sparse points across entire videos. PIPs~\cite{pips} introduced a deep tracking framework for point tracking, followed by TAP-Net~\cite{tapvid}, TAPIR~\cite{tapir}, and CoTracker~\cite{cotracker}, which rely on correlation-based matching and iterative updates to propagate tracks over time. These approaches operate purely in 2D and typically depend on carefully designed matching and update mechanisms. 
Recent 3D point tracking approaches extend this paradigm by decoupling geometry reconstruction from motion modeling. 
SpatialTracker~\cite{xiao2024spatialtracker}, and subsequent methods~\cite{ngo2024delta, xiao2025spatialtracker, tapip3d} combine a pre-trained depth estimator with a lifted 2D tracking pipeline~\cite{cotracker} to operate in 3D. Despite enabling 3D tracking, their multi-stage pipelines remain limited in efficiency and flexibility, and they do not learn a unified spatiotemporal representation. 
In contrast, \nickname directly models dense geometry and motion jointly within a unified feed-forward framework, without decoupled stages or tracking heuristics.


\noindent
\textbf{4D Reconstruction.}
The goal of 4D reconstruction is to recover a representation that captures both the 3D structure of a scene and how it evolves over time. Early methods~\cite{wang2023omnimotion, som, lei2024mosca, wang2025gflow} typically formulate this problem as test-time optimization, which can produce high-quality results but requires costly per-scene optimization. Recent efforts have gradually shifted toward feed-forward formulations of 4D reconstruction. St4RTrack~\cite{st4rtrack2025} predicts point maps for pairs of views, jointly encoding static geometry and dynamic motion; however, its pairwise formulation inherently limits the temporal range of the reconstruction. 
We also acknowledge several recent concurrent works that explore feed-forward formulations for 4D reconstruction.
TraceAnything~\cite{liu2025traceanythingrepresentingvideo} represents scenes using continuous trajectory fields parameterized by Bézier curves. Although this formulation enables smooth and long-range motion modeling, it often struggles to represent complex or high-frequency dynamics and may compromise geometric accuracy. 
Any4D~\cite{karhade2025any4d} jointly predicts scene flow and 3D geometry from a canonical reference view, but lacks the flexibility to infer motion originating from arbitrary viewpoints. 
Similarly, V-DPM~\cite{sucar2025vdpm} extends VGGT to dynamic settings, but relies on an inflexible decoding scheme that aggregates information from all views, leading to high computational costs.
%
%
Concurrently, D4RT~\cite{zhang2025d4rt} introduces a unified model for 2D and 3D point tracking. 
Specifically, D4RT first encodes the entire video into a global scene representation using a self-attention encoder, and then answers spatio-temporal per-point queries through a lightweight cross-attention decoder.
%
Likewise, our method, \nickname, employs a flexible query-based decoder that efficiently recovers complete and dense 4D attributes for any view at any timestamp, without expensive per-point computation.

\vspace{2mm}
\section{Method}
\label{sec:method}
\vspace{1mm}

\begin{figure*}[t]
    \begin{center}
\centerline{\includegraphics[width=\textwidth]{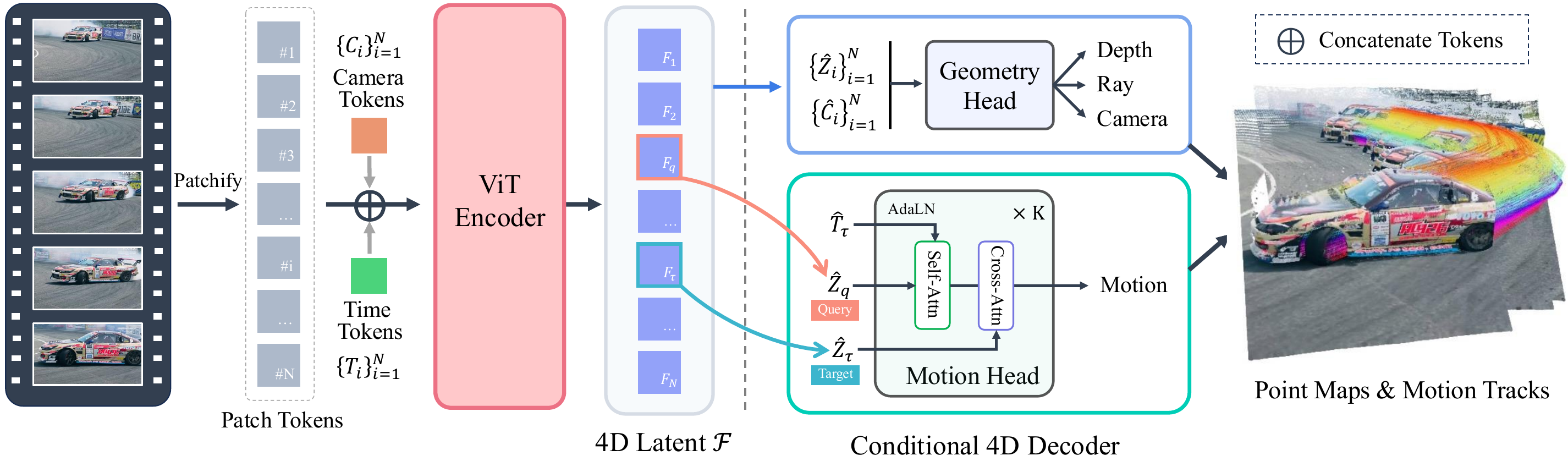}}
    \caption{\textbf{Overall architecture of \nickname.} 
    Video frames are patchified and augmented with camera and time tokens, then jointly encoded by a single transformer into a compact 4D latent representation $\mathcal{F}$, from which a conditional decoder with disentangled geometry and motion heads enables flexible querying of 3D geometry and motion for arbitrary source views at arbitrary target timestamps.
    }
    \label{fig:architecture}
    \end{center}
    \vspace{-5mm}
\end{figure*}

Our goal is to develop a unified and feed-forward model, \nickname, that takes a monocular video as input and reconstructs the full underlying 4D attributes of the scene. The core of our approach lies in encoding the entire video sequence into a compact 4D representation, which can then be queried on-demand to decode the geometry and motion of any query frame at any target timestamp, as illustrated in Figure~\ref{fig:architecture}.

\subsection{Problem Formulation}

Given a monocular video sequence $\mathcal{V} = \{ I_i \}_{i=1}^N$, where $I_i \in \mathbb{R}^{H \times W \times 3}$ denotes the RGB frame captured at timestamp $t_i$ and $N$ is the total number of frames, our goal is to recover the full 4D attributes of the scene, capturing both its 3D structure and temporal evolution.
Specifically, for any query frame $I_i$ and an arbitrary target timestamp $\tau \in \{t_i\}_{i=1}^N$, we define a time-indexed 3D point map:
\begin{equation}
P_i^{t_i \rightarrow \tau} \in \mathbb{R}^{H \times W \times 3},
\end{equation}
which represents the 3D positions of points observed in frame $I_i$ as they appear at time $\tau$.
When $\tau = t_i$, $P_i^{t_i \rightarrow \tau}$ corresponds to the static 3D geometry of the frame. 
When $\tau \neq t_i$, it describes the dynamic time-dependent point maps of the scene by mapping the points from the source frame to their locations at the target time.

\noindent {\bf Factorized 4D Attributes.}
Directly predicting point maps $P_i^{t_i \rightarrow \tau}$ for all possible $(i, \tau)$ pairs is redundant and intractable. 
Once the underlying 3D geometry at the source time is known, the geometry at other times can be expressed through relative motion. We therefore adopt a factorized representation:
\begin{equation}
    P_i^{t_i \rightarrow \tau} = P_i^{t_i} + \Delta P_i^{t_i \rightarrow \tau},
\end{equation}
where $P_i^{t_i}$ denotes the base 3D geometry at time $t_i$, and $\Delta P_i^{t_i \rightarrow \tau}$ represents the 3D displacement from time $t_i$ to $\tau$.

This formulation offers both \textit{conceptual} and \textit{practical} advantages. The base geometry $P_i^{t_i}$ is reconstructed from image $I_i$ under the perspective camera model, a property that allows us to directly leverage recent advances of effective geometry representation in monocular 3D reconstruction~\cite{depthanything3}.
Meanwhile, the displacement field $\Delta P_i^{t_i \rightarrow \tau}$ explicitly captures temporal motion. This provides clear motion cues that are useful for downstream applications, while avoiding the need to re-predict complex geometry at every time step. As a result, the representation remains temporally consistent, especially in static regions and under rigid motion.
Unless otherwise stated, all point maps are viewpoint-invariant and expressed in a world coordinate system defined by the camera of the first frame~\cite{wang2024dust3r, wang2025cut3r, wang2025vggt, depthanything3}.

\noindent {\bf Relation with Other Work.}
The key distinction between \nickname and several prior or concurrent approaches lies in the flexibility and completeness of our 4D output. 
Recent feed-forward 3D reconstruction methods focus solely on predicting the base 3D geometry for each input frame, i.e., $P_i^{t_i}$, and thus fail to capture the motion within the scene.
Traditional 3D point tracking methods, on the other hand, estimate sparse trajectories initialized from selected points and therefore cannot recover dense 4D geometry. Concurrent feed-forward 4D reconstruction methods also exhibit limitations in motion modeling. 
St4RTrack is restricted to pairwise motion. TraceAnything models trajectory fields using Bézier curves, which limits its ability to capture accurate geometry and complex motion. Any4D predicts motion only relative to the first frame, i.e., $P_1^{t_1 \rightarrow \tau}$ with $\tau \in \{t_i\}_{i=1}^N$, and therefore cannot support motion queries from other source frames. V-DPM regresses the point map $P_i^{t_i \rightarrow \tau}$ for all source frames
$i \in \{1, \dots, N\}$ at a given target timestamp $\tau$,
by attending to all frames jointly, which incurs substantial computational overhead and limits inference flexibility. In contrast, \nickname enables flexibly querying dense 3D motion from any single source frame to any target timestamp within a unified and fully feed-forward framework.

\subsection{4D Representation Encoder}
The encoder $\mathcal{E}$ processes the input video $\mathcal{V}$ to produce a unified 4D representation:
\begin{equation}
\mathcal{F} = \mathcal{E}(\mathcal{V}).
\end{equation}
We adopt a plain ViT-based transformer architecture that alternates between frame-wise self-attention and global self-attention.
Similar to the camera token in VGGT~\cite{wang2025vggt}, which primarily encodes camera geometry information for subsequent decoding, we further append each view’s patchified tokens with a dedicated time token $T_i$. This time token aggregates temporal information for that view and serves as a conditioning signal for target-time motion decoding, as described in Section~\ref{sec:method_decoder}.
The encoder produces a unified spatio-temporal latent representation
$\mathcal{F} = \{F_i\}_{i=1}^N$.
Each $F_i = \{\hat{Z}_{i,j}\}_{j=1}^{M} \cup \{~\hat{C}_i\} \cup \{\hat{T}_i\}$ consists of $M$ patch tokens $\hat{Z}_{i,j} \in \mathbb{R}^D$ corresponding to the $i$-th frame, together with an encoded camera token $\hat{C}_i$ and a time token $\hat{T}_i$.
We treat $\mathcal{F}$ as an ordered sequence of frame-level token sets.

\subsection{Conditional 4D Decoder}
\label{sec:method_decoder}

\noindent {\bf Geometry Head.}
To recover the base geometry for each input frame, we use a geometry decoder $\mathcal{D}_{\mathrm{g}}$.
Given the encoded spatial tokens $\hat{Z}_i$ and camera tokens $\hat{C}_i$, the geometry decoder predicts per-frame depth and rays, together with camera parameters:
\begin{equation}
\left(\hat{D}_i,\, \hat{R}_i,\, \hat{\theta}_i\right) = \mathcal{D}_{\mathrm{g}}\!\left(\hat{Z}_i, \hat{C}_i\right),
\end{equation}
where $\hat{D}_i \in \mathbb{R}^{H \times W}$ is the depth map, $\hat{R}_i \in \mathbb{R}^{\frac{1}{2}H \times \frac{1}{2}W \times 6}$ is the ray map, and $\hat{\theta}_i$ denotes the camera parameters (i.e., field of view, rotation, and translation).
The base point map $P_i^{t_i}$ is then obtained from $(\hat{D}_i, \hat{R}_i, \theta_i)$ under the perspective camera model.
The geometry decoder $\mathcal{D}_{\mathrm{g}}$ follows a dual-DPT~\cite{dpt,depthanything3} design with a lightweight camera head.

\noindent {\bf Motion Head.}  
\label{sec:method_decoder:motion_head}
To recover motion for any query frame $I_q$ at a target timestamp $\tau$, we use a lightweight transformer-based motion decoder $\mathcal{D}_{\mathrm{m}}$ with $K$ layers of alternating self-attention and cross-attention.  
We initialize the query tokens $\hat{Z}_q$ from the encoder output $\mathcal{F}$.  
The decoder outputs a dense 3D displacement field:
\begin{equation}
\Delta \hat{P}_q^{t_q \rightarrow \tau} = \mathcal{D}_{\mathrm{m}}\!\left(\hat{Z}_q,\, \hat{T}_{\tau},\, \hat{Z}_{\tau}\right).
\end{equation}
Specifically, to condition on the target time, we inject time embedding $\hat{T}_{\tau}$ via Adaptive Layer Normalization (AdaLN)~\cite{perez2018film} in the self-attention blocks, and then apply cross-attention to the target spatial token set $\hat{Z}_{\tau}$. This design supports dense motion estimation and point tracking while remaining compatible with our per-frame geometry decoding.

\subsection{Training Scheme}
\label{subsec:training}

We train \nickname in an end-to-end manner with joint supervision over geometry and motion attributes. Following prior works~\cite{wang2025vggt, depthanything3}, we normalize the ground-truth scene scale such that the average Euclidean distance of all valid 3D points to the origin is 1. The overall training objective is defined as:
\begin{equation}
    \mathcal{L} =
    \mathcal{L}_{\text{depth}} +
    \mathcal{L}_{\text{ray}} +
    \mathcal{L}_{\text{cam}} +
    \mathcal{L}_{\text{motion}}.
    \label{eq:loss}
\end{equation}
\noindent
For all loss terms except the camera parameter loss $\mathcal{L}_{\text{cam}}$, we adopt an aleatoric uncertainty formulation~\cite{wang2024dust3r}.  We denote the loss function as
$\ell(\hat{\mathbf{y}}, \mathbf{y}, \mathbf{\Sigma})$,
where $\mathbf{\Sigma}$ represents the predicted pixel-wise uncertainty map, which adaptively down-weights unreliable regions during training.

To better supervise both geometry and motion, we apply gradient-based constraints~\cite{depthanything3} in the spatial and temporal domains separately. For geometry learning, we enforce spatial smoothness on the predicted depth maps $\hat{\mathbf{D}} = \{ \hat{D}_i \}$ by applying image-space gradients $\nabla_{\mathbf{x}}$. The depth loss is formulated as:
\begin{equation}
    \mathcal{L}_{\text{depth}} =
    \ell(\hat{\mathbf{D}}, \mathbf{D}, \mathbf{\Sigma}_D)
    +
    \ell(\nabla_{\mathbf{x}}\hat{\mathbf{D}}, \nabla_{\mathbf{x}}\mathbf{D}, \mathbf{\Sigma}_D).
\end{equation}

Similarly, the motion loss supervises the displacement field $\Delta \mathbf{P}$, but we incorporate an additional temporal gradient term $\nabla_t$ that constrains the first-order temporal derivative of the displacement (i.e., velocity) to encourage temporally consistent motion behavior:
%
\begin{equation}
    \begin{split}
        \mathcal{L}_{\text{motion}} = \ell(\Delta \hat{\mathbf{P}}, \Delta \mathbf{P}, \mathbf{\Sigma}_M) \\
        +
        \ell(\nabla_t \Delta \hat{\mathbf{P}}, \nabla_t \Delta \mathbf{P}, \mathbf{\Sigma}_M).
    \end{split}
\end{equation}

\begin{figure*}[!t]
    \begin{center}
    \centerline{\includegraphics[width=\textwidth]{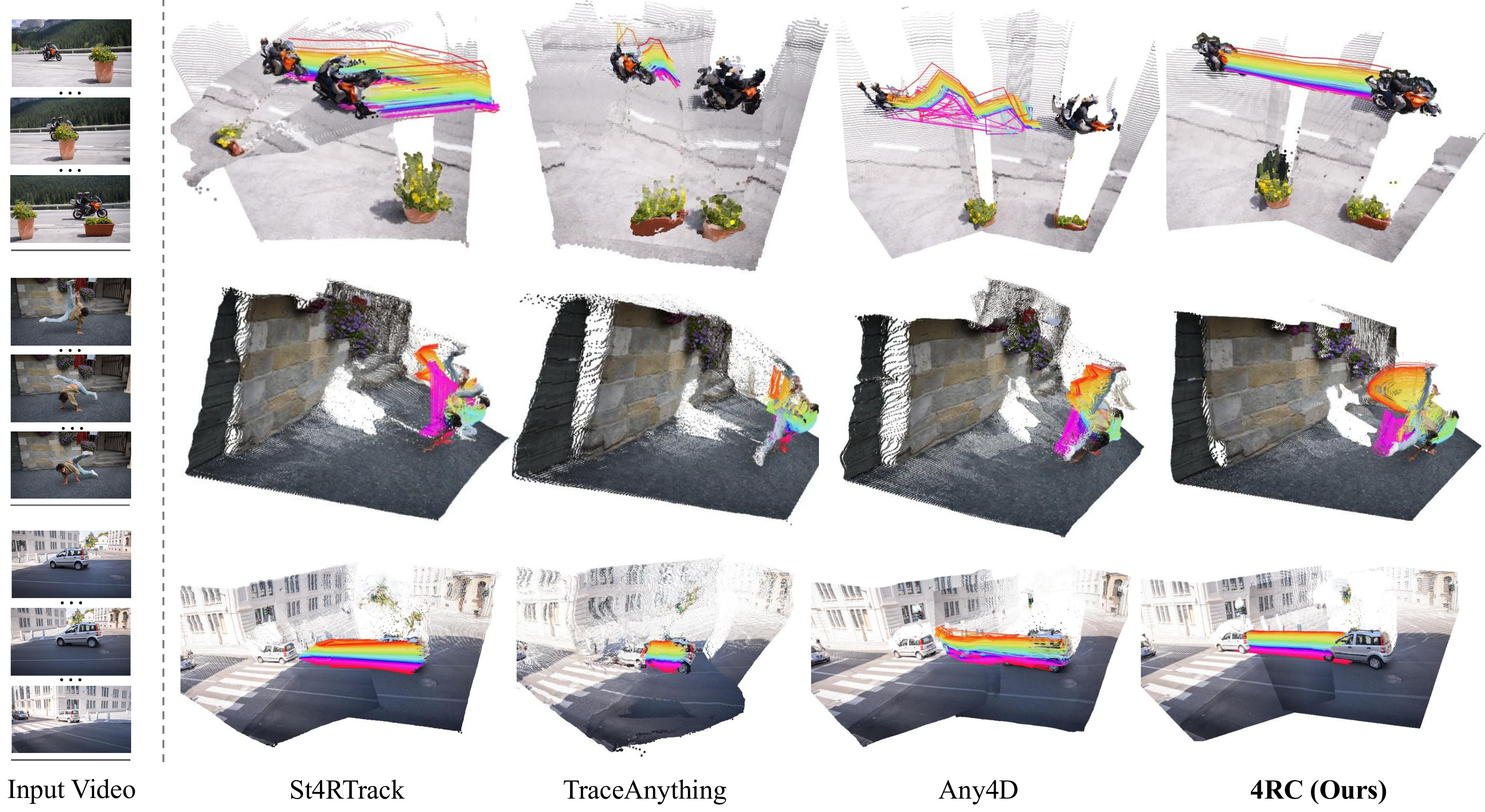}}
    \caption{\textbf{Qualitative comparison of dynamic tracking on DAVIS}~\cite{davis_CVPR_2016}\textbf{.} We visualize the dynamic reconstruction results, including the geometry at the first and last frames, as well as the dynamic object trajectories rendered as rainbow-colored paths from the first view. As shown in the top example, our method successfully handles occlusion when the motorcycle becomes temporarily invisible. In contrast, the two-view method St4RTrack lacks global temporal context and therefore predicts an incorrect trajectory. In the second and third examples, our method accurately reconstructs complex and large-scale motions while preserving high-quality geometry, while other methods produce inconsistent motion trajectories and degraded geometry.}
    \label{fig:qualitative_main}
    \end{center}
    \vspace{-5mm}
\end{figure*}

\section{Experiments}
\label{sec:experiments}
We conduct extensive experiments to evaluate the effectiveness of \nickname on standard 4D reconstruction tasks. We compare against established state-of-the-art methods as well as concurrent work for completeness, and further perform ablation studies to analyze the contribution of key design components in our framework.

\vspace{-2mm}

\subsection{Training Setup}

\noindent {\bf Datasets.}
We train \nickname on a diverse collection of large-scale public datasets, covering both dynamic and static scenes, as well as synthetic and real-world videos. Specifically, our training data includes PointOdyssey~\cite{zheng2023point}, Dynamic Replica~\cite{karaev2023dynamicstereo}, Kubric~\cite{greff2021kubric}, Waymo~\cite{waymo}, DL3DV~\cite{ling2024dl3dv}, ScanNet++~\cite{yeshwanth2023scannet++}, and MVS-Synth~\cite{DeepMVS}. These datasets jointly provide rich supervision for geometry, motion, and camera poses under varied scene layouts and motion patterns. Detailed dataset statistics are provided in the appendix.

\noindent {\bf Implementation Details.}
Our encoder adopts a single Vision Transformer based on DINOv2~\cite{oquab2023dinov2}. The motion decoder is lightweight, consisting of $K = 4$ layers of self-attention and cross-attention.
We initialize both the encoder and the geometry decoder with pretrained weights from DA3~\cite{depthanything3}, which is trained on large-scale 3D data and provides strong geometric priors. During training, input images are resized to a randomly sampled resolution, with the longer side up to 504 pixels. The aspect ratio is uniformly sampled from $[0.5, 2.0]$ to improve generalizability. The training sequence length $N$ is randomly sampled from [2, 18] views, with longer sequences facilitating larger and more complex motions. To avoid the quadratic cost of computing all $N^2$ motion pairs, we randomly sample one query view per iteration and predict its motion in $N$ different timesteps during training. Standard data augmentations including color jittering and Gaussian blur are applied. The model is trained end-to-end using the training loss described in Section~\ref{subsec:training}. We use the AdamW optimizer~\cite{Kingma2015AdamAM, loshchilov2018decoupled} for 50 epochs with a cosine learning rate schedule. Training is performed on 16 A100 GPUs with a batch size of 1 per GPU. Additional implementation details and hyperparameters are provided in the appendix.

\vspace{-2mm}
\subsection{4D Reconstruction}

\begin{figure}[!t]
\begin{center}
\centerline{\includegraphics[width=\columnwidth]{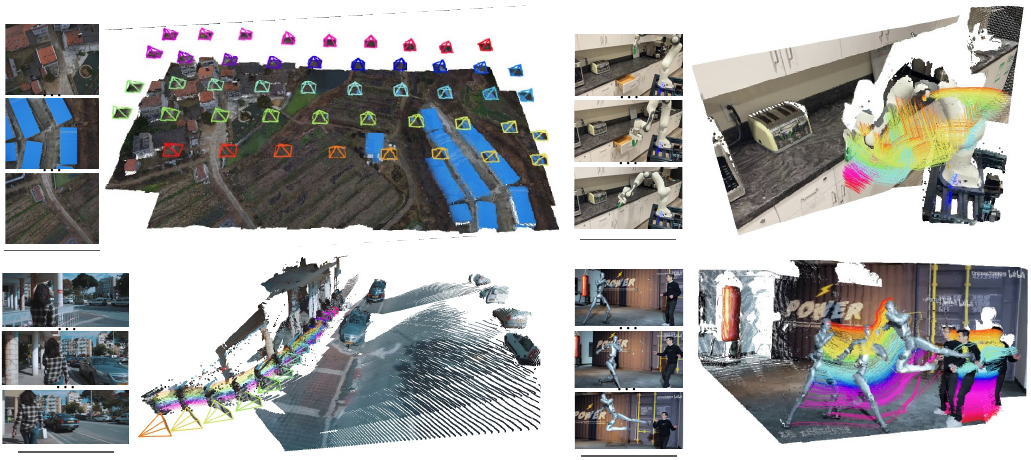}}
\caption{\textbf{Visualization of in-the-wild examples}. \nickname demonstrates accurate geometry reconstruction and motion modeling in both static and dynamic scenes.}
\label{fig:qualitive_itw}
\end{center}
\vspace{-10mm}
\end{figure}

\noindent {\bf Qualitative Results.} 
Figure~\ref{fig:qualitative_main} provides qualitative comparisons of \nickname in modeling 3D tracking. These visual results demonstrate the effectiveness of our method in handling complex motion patterns, such as occlusions, non-rigid motion, and large movements. We further evaluate our method on diverse in-the-wild videos in Figure~\ref{fig:qualitive_itw}, demonstrating its strong performance on both static and dynamic scenes.

\definecolor{best}{RGB}{180, 220, 200}
\definecolor{second}{RGB}{225, 250, 240}

\begin{table*}[!t]
\centering
\caption{\textbf{4D reconstruction evaluation on tracking}. We evaluate our method on dense-view tracking (a), as well as sparse-view tracking (b) on dynamic datasets. Our method demonstrates state-of-the-art capability in dense tracking from arbitrary views compared to concurrent 4D reconstruction methods, and also achieves strong performance on the sparse point tracking setting, even when compared to tracking-specific methods. The top-2 results are highlighted as \colorbox{best}{best} and \colorbox{second}{second}.}

\label{tab:world_tracking_merged}
\renewcommand{\arraystretch}{1.15}
\renewcommand{\tabcolsep}{1.2mm}
\resizebox{0.98\textwidth}{!}{%
\begin{tabular}{l cc cc cc cc cc cc}
\toprule
\multirow{3}{*}{Method} & \multicolumn{4}{c}{(a) Dense Tracking} & \multicolumn{8}{c}{(b) Sparse Point Tracking} \\
\cmidrule(lr){2-5} \cmidrule(lr){6-13}
  & \multicolumn{2}{c}{Kubric} & \multicolumn{2}{c}{Waymo} & \multicolumn{2}{c}{PO} & \multicolumn{2}{c}{DR} & \multicolumn{2}{c}{ADT} & \multicolumn{2}{c}{PStudio} \\
\cmidrule(lr){2-3} \cmidrule(lr){4-5} \cmidrule(lr){6-7} \cmidrule(lr){8-9} \cmidrule(lr){10-11} \cmidrule(lr){12-13} 
  & APD $\uparrow$ & EPE $\downarrow$ & APD $\uparrow$ & EPE $\downarrow$ & APD $\uparrow$ & EPE $\downarrow$ & APD $\uparrow$ & EPE $\downarrow$ & APD $\uparrow$ & EPE $\downarrow$ & APD $\uparrow$ & EPE $\downarrow$ \\
\midrule
VGGT + CoTracker3~\cite{cotracker3} & - & - & - & - & 63.19 & 0.5890 & 80.93 & 0.2417 & 77.81 & 0.3015 & 78.11 & 0.2715 \\
SpatialTrackerV2~\cite{xiao2025spatialtracker} & - & - & - & - & 73.66 & 0.3944 & 80.87 & 0.2218 & \cellcolor{best}95.48 & \cellcolor{best}0.0594 & 85.63 & 0.1583 \\
St4RTrack~\cite{st4rtrack2025}        & 50.65 & 3.938 & 19.98 & 6.359 & 71.64 & 0.3101 & 78.36 & 0.2367 & 82.79 & 0.2279 & 74.05 & 0.2537 \\
TraceAnything~\cite{liu2025traceanythingrepresentingvideo} & 59.98 & \cellcolor{second}1.808 & 21.25 & 4.313 & 52.02 & 0.9154 & 68.28 & 0.5060 & 82.77 & 0.1998 & 74.15 & 0.2926 \\
Any4D~\cite{karhade2025any4d}        & - & - & - & - & 71.47 & 0.3642 & 81.28 & 0.2171 & 73.83 & 0.3114 & 78.76 & 0.2088 \\
V-DPM~\cite{sucar2025vdpm}        & \cellcolor{second}71.12 & 2.849 & \cellcolor{second}41.44 & \cellcolor{second}1.948 & \cellcolor{second}83.36 & \cellcolor{best}0.1955 & \cellcolor{second}83.04 & \cellcolor{second}0.1901 & 80.80 & 0.2357 & \cellcolor{best}89.59 & \cellcolor{best}0.1165 \\
\midrule
\textbf{\nickname~(Ours)} & \cellcolor{best}85.44 & \cellcolor{best}1.022 & \cellcolor{best}56.63 & \cellcolor{best}1.611 & \cellcolor{best}85.86 & \cellcolor{second}0.2498 & \cellcolor{best}88.65 & \cellcolor{best}0.1484 & \cellcolor{second}87.82 & \cellcolor{second}0.1480 & \cellcolor{second}87.32 & \cellcolor{second}0.1304 \\
\bottomrule
\end{tabular}
}
\vspace{-1.5mm}
\end{table*}

\noindent {\bf Dense Tracking.}
To demonstrate the capability of our method to track dense motion from arbitrary query views, we first quantitatively evaluate dense 3D tracking by sampling 24 frames from the Kubric and Waymo test sets, with 50 samples each, using the middle view (i.e., the 11th frame) as the query. Traditional point tracking methods fail on dense tracking due to out-of-memory issues, while Any4D can only predict the motion field for the first view. We report both the \textit{Average Percentage of Points} (APD) within a threshold and the \textit{End-Point Error} (EPE) after global Sim(3) alignment with RANSAC. As shown in Table~\ref{tab:world_tracking_merged}(a), \nickname achieves state-of-the-art performance among concurrent 4D reconstruction methods on both datasets. On the challenging Waymo dataset, which contains highly dynamic scenes, our method substantially outperforms the concurrent method V-DPM, resulting in a 36\% gain in APD. Notably, our method uses flexible per-frame decoding, in contrast to V-DPM’s computationally expensive global aggregation decoding.

\noindent {\bf Sparse Point Tracking.}
We then evaluate \nickname on 3D sparse point tracking, which measures sparse motion relative to the first frame, although our method can fully capture dense motion. Following the WorldTrack benchmark~\cite{st4rtrack2025}, tracking performance is assessed in the world coordinate system. The benchmark includes two datasets, Aerial Digital Twin (ADT)~\cite{pan2023ariadigitaltwinnew} and Panoptic Studio (PStudio)~\cite{pstudio} from TAPVid-3D~\cite{koppula2024tapvid3d}, as well as two test sets from PointOdyssey (PO) and Dynamic Replica (DR). We compare our method against tracking-specific methods Cotracker3~\cite{cotracker3} and SpatialTrackerV2~\cite{xiao2025spatialtracker}, along with concurrent 4D reconstruction methods. The predicted trajectory is aligned to the ground truth using a global Sim(3) transformation via RANSAC. As shown in Table~\ref{tab:world_tracking_merged}(b), \nickname achieves strong performance even when compared with methods specifically designed for point tracking, outperforming SpatialTrackerV2 on 3 out of 4 datasets.



\definecolor{best}{RGB}{180, 220, 200}
\definecolor{second}{RGB}{225, 250, 240}

\begin{table*}[t]
\centering
\caption{\textbf{Camera pose estimation and multi-View 3D reconstruction evaluation.} We compare our method with both 3D reconstruction approaches and concurrent 4D reconstruction methods. Our approach achieves state-of-the-art performance among 4D methods, while remaining competitive with 3D reconstruction methods Pi3, without exclusive training on large-scale reconstruction datasets.}

\label{tab:pose_recon_nosintel_swapped_nobold}
\renewcommand{\arraystretch}{1.15}
\renewcommand{\tabcolsep}{1.5mm}
\resizebox{0.98\textwidth}{!}{%
\begin{tabular}{l ccc ccc ccc ccc}
\toprule
\multirow{3}{*}{Method} & \multicolumn{6}{c}{(a) Camera Pose Estimation} & \multicolumn{6}{c}{(b) Multi-View 3D Reconstruction} \\
\cmidrule(lr){2-7} \cmidrule(lr){8-13}
 & \multicolumn{3}{c}{TUM-dynamics} & \multicolumn{3}{c}{ScanNet} & \multicolumn{3}{c}{7-Scenes} & \multicolumn{3}{c}{NRGBD} \\
\cmidrule(lr){2-4} \cmidrule(lr){5-7} \cmidrule(lr){8-10} \cmidrule(lr){11-13}
 & ATE $\downarrow$ & RPE$_\text{t}$ $\downarrow$ & RPE$_\text{r}$ $\downarrow$ & ATE $\downarrow$ & RPE$_\text{t}$ $\downarrow$ & RPE$_\text{r}$ $\downarrow$ & Acc $\downarrow$ & Comp $\downarrow$ & NC $\uparrow$ & Acc $\downarrow$ & Comp $\downarrow$ & NC $\uparrow$ \\
\midrule
DUSt3R~\cite{wang2024dust3r} & 0.083 & 0.017 & 3.567 & 0.081 & 0.028 & 0.784 & 0.146 & 0.181 & 0.736 & 0.144 & 0.154 & 0.870 \\
MASt3R~\cite{mast3r_arxiv24} & 0.038 & 0.012 & 0.448 & 0.078 & 0.020 & 0.475 & 0.185 & 0.180 & 0.701 & 0.085 & 0.063 & 0.794 \\
MonST3R~\cite{zhang2024monst3r} & 0.098 & 0.019 & 0.935 & 0.077 & 0.018 & 0.529 & 0.248 & 0.266 & 0.672 & 0.272 & 0.287 & 0.758 \\
Spann3R~\cite{wang2024spann3r} & 0.056 & 0.021 & 0.591 & 0.096 & 0.023 & 0.661 & 0.298 & 0.205 & 0.650 & 0.416 & 0.417 & 0.684 \\
CUT3R~\cite{wang2025cut3r} & 0.046 & 0.015 & 0.473 & 0.099 & 0.022 & 0.600 & 0.126 & 0.154 & 0.727 & 0.099 & 0.076 & 0.837 \\
VGGT~\cite{wang2025vggt} & \cellcolor{second}0.012 & 0.010 & \cellcolor{second}0.311 & 0.036 & 0.015 & \cellcolor{second}0.376 & 0.087 & 0.091 & \cellcolor{best}0.787 & 0.073 & 0.077 & 0.910 \\
Pi3~\cite{wang2025pi} & 0.014 & \cellcolor{second}0.009 & \cellcolor{best}0.309 & \cellcolor{best}0.031 & \cellcolor{second}0.013 & \cellcolor{best}0.346 & \cellcolor{second}0.044 & \cellcolor{second}0.063 & 0.758 & \cellcolor{best}0.022 & \cellcolor{best}0.025 & \cellcolor{second}0.911 \\
St4RTrack~\cite{st4rtrack2025} & - & - & - & - & - & - & 0.240 & 0.234 & 0.681 & 0.241 & 0.219 & 0.754 \\
TraceAnything~\cite{liu2025traceanythingrepresentingvideo} & - & - & - & - & - & - & 0.232 & 0.359 & 0.584 & 0.347 & 0.527 & 0.643 \\
Any4D~\cite{karhade2025any4d} & 0.030 & 0.023 & 0.463 & 0.074 & 0.035 & 1.076 & 0.141 & 0.177 & 0.738 & 0.081 & 0.072 & 0.847 \\
V-DPM~\cite{sucar2025vdpm} & 0.014 & 0.010 & 0.318 & 0.035 & 0.014 & 0.410 & 0.097 & 0.124 & 0.772 & 0.056 & 0.060 & 0.897 \\
\midrule
\textbf{\nickname~(Ours)} & \cellcolor{best}0.010 & \cellcolor{best}0.008 & 0.314 & \cellcolor{second}0.032 & \cellcolor{best}0.012 & 0.437 & \cellcolor{best}0.034 & \cellcolor{best}0.051 & \cellcolor{second}0.783 & \cellcolor{second}0.036 & \cellcolor{second}0.034 & \cellcolor{best}0.912 \\
\bottomrule
\end{tabular}
}
\vspace{-1mm}
\end{table*}

\subsection{3D Reconstruction}

\noindent {\bf Camera Pose Estimation.}  
We evaluate camera pose estimation on the Sintel~\cite{sintel}, TUM-dynamics~\cite{tum-dynamics}, and ScanNet~\cite{dai2017scannet} datasets. Performance is measured using \textit{Absolute Translation Error} (ATE), \textit{Relative Translation Error} (RPE$_\text{t}$), and \textit{Relative Rotation Error} (RPE$_\text{r}$), all computed after global Sim(3) alignment with the ground truth, following established protocols~\cite{teed2021droid, zhang2024monst3r, wang2025cut3r}. Table~\ref{tab:pose_recon_nosintel_swapped_nobold}~(a) shows that \nickname achieves top-tier camera pose estimation and reconstruction quality within a single unified model. On the challenging TUM-dynamics dataset, \nickname attains the best ATE and RPE$_\text{t}$ among all methods, including specialized 3D reconstruction methods such as Pi3, which are trained on much larger datasets. This demonstrates that our unified 4D representation is effective for both motion modeling and producing accurate camera trajectories. Notably, \nickname achieves the best performance among concurrent feed-forward 4D reconstruction methods. We exclude St4RTrack and TraceAnything as they do not explicitly estimate camera poses.

\noindent {\bf Multi-View Reconstruction.}  
Following prior work~\cite{wang2024spann3r, wang2025cut3r, wang2024dust3r}, we evaluate scene-level multi-view 3D reconstruction on the 7-Scenes~\cite{Shotton_2013_CVPR} and NRGBD~\cite{Azinovic_2022_CVPR} datasets. Reconstruction quality is measured using \textit{Accuracy} (Acc), \textit{Completeness} (Comp), and \textit{Normal Consistency} (NC). Quantitative results are reported in Table~\ref{tab:pose_recon_nosintel_swapped_nobold}~(b). \nickname achieves the best performance among 4D reconstruction methods, attaining the highest Acc/Comp on 7-Scenes and the best NC on NRGBD. This highlights the effectiveness of our proposed design.
For example, we obtain 0.034 accuracy on 7-Scenes, far better than TraceAnything’s 0.240; the latter jointly models geometry and motion in a trajectory field, which often compromises geometric quality.

\definecolor{best}{RGB}{180, 220, 200}
\definecolor{second}{RGB}{225, 250, 240}
\begin{table}[!t]
\centering
\caption{\textbf{Depth estimation on the Bonn and Sintel datasets.} We compare methods that explicitly predict video depth.}
\label{tab:depth_estimation_swapped}
\renewcommand{\arraystretch}{1.15}
\renewcommand{\tabcolsep}{0.6mm}
\resizebox{0.98\columnwidth}{!}{
\begin{tabular}{lcc cc}
\toprule
\multirow{2}{*}{Method} & \multicolumn{2}{c}{Bonn} & \multicolumn{2}{c}{Sintel} \\
\cmidrule(lr){2-3} \cmidrule(lr){4-5}
 & Rel $\downarrow$ & $\delta < 1.25 \uparrow$ & Rel $\downarrow$ & $\delta < 1.25 \uparrow$ \\
\midrule
DUSt3R~\cite{wang2024dust3r} & 0.155 & 83.3 & 0.656 & 45.2 \\
MASt3R~\cite{mast3r_arxiv24} & 0.252 & 70.1 & 0.641 & 43.9 \\
MonST3R~\cite{zhang2024monst3r} & 0.067 & 96.3 & 0.378 & 55.8 \\
Spann3R~\cite{wang2024spann3r} & 0.144 & 81.3 & 0.622 & 42.6 \\
CUT3R~\cite{wang2025cut3r} & 0.078 & 93.7 & 0.421 & 47.9 \\
Fast3R~\cite{Yang_2025_Fast3R} & 0.193 & 77.5 & 0.653 & 44.9 \\
VGGT~\cite{wang2025vggt} & 0.055 & 97.1 & \cellcolor{second}0.297 & \cellcolor{best}68.8 \\
Pi3~\cite{wang2025pi} & \cellcolor{best}0.050 & \cellcolor{best}97.4 & \cellcolor{best}0.246 & \cellcolor{second}67.7 \\
\midrule
\textbf{\nickname~(Ours)} & \cellcolor{second}0.051 & \cellcolor{best}97.4 & 0.311 & 62.2 \\
\bottomrule
\end{tabular}
}
\vspace{-3mm}
\end{table}

\noindent {\bf Depth Estimation.}  
We also evaluate video depth estimation on Sintel~\cite{sintel} and Bonn~\cite{Bonn} datasets. Following prior work~\cite{wang2025cut3r}, predicted depth maps are aligned to the ground truth using a per-sequence scale factor. 
While most existing 4D reconstruction methods do not explicitly output depth and therefore cannot be directly evaluated on depth benchmarks, \nickname includes an explicit depth prediction as part of its factorized 4D representation.
On the Bonn dataset, \nickname achieves the best $\delta < 1.25$ score and matches the second-best Rel. On Sintel, there is a small gap compared to specialized 3D reconstruction methods such as Pi3, which are trained exclusively on large-scale 3D datasets that are more than twice the size of our training datasets.

\subsection{Ablation Studies}

\definecolor{graybg}{gray}{0.95}

\begin{table}[t]
\centering
\caption{\textbf{Ablation of our motion head design and factorized motion.} In (a), we evaluate the effectiveness of our motion head design by removing each component. (b) shows that representing the motion output in a factorized form performs better than directly predicting the point cloud.}
\label{tab:ablation}
\renewcommand{\arraystretch}{1.1}
\renewcommand{\tabcolsep}{3.5mm}
\resizebox{\linewidth}{!}{
\begin{tabular}{l cc cc}
\toprule
\multirow{2}{*}{Methods} & \multicolumn{2}{c}{Kubric} & \multicolumn{2}{c}{Waymo} \\
\cmidrule(lr){2-3} \cmidrule(lr){4-5}
 & APD $\uparrow$ & EPE $\downarrow$ & APD $\uparrow$ & EPE $\downarrow$ \\
\midrule

\textbf{4RC (Ours)} & \textbf{85.44} & \textbf{1.022} & \textbf{56.63} & \textbf{1.611} \\

\midrule
\rowcolor{graybg} \multicolumn{5}{c}{\textit{(a) Motion Head Design}} \\
\midrule
(i)~~~w/o Cross Attn.   & 80.83 & 1.136 & 54.19 & 1.618 \\
(ii)~~w/o Self Attn.     & 80.57 & 1.127 & 53.50 & 1.686 \\
(iii) w/o AdaLN          & 82.51 & 1.105 & 56.11 & 1.689 \\

\midrule
\rowcolor{graybg} \multicolumn{5}{c}{\textit{(b) Factorized Motion}} \\
\midrule
(i)~~Points (World)    & 74.64 & 1.412 & 37.08 & 2.359 \\
(ii)~Points (Local)     & 70.70 & 1.547 & 19.55 & 3.226 \\

\bottomrule
\end{tabular}
}
\vspace{-5mm}
\end{table}

We conduct ablation studies to evaluate the key design choices in \nickname, focusing on the motion head and the factorized motion representation.

\begin{figure}[t]
\begin{center}
\centerline{\includegraphics[width=\columnwidth]{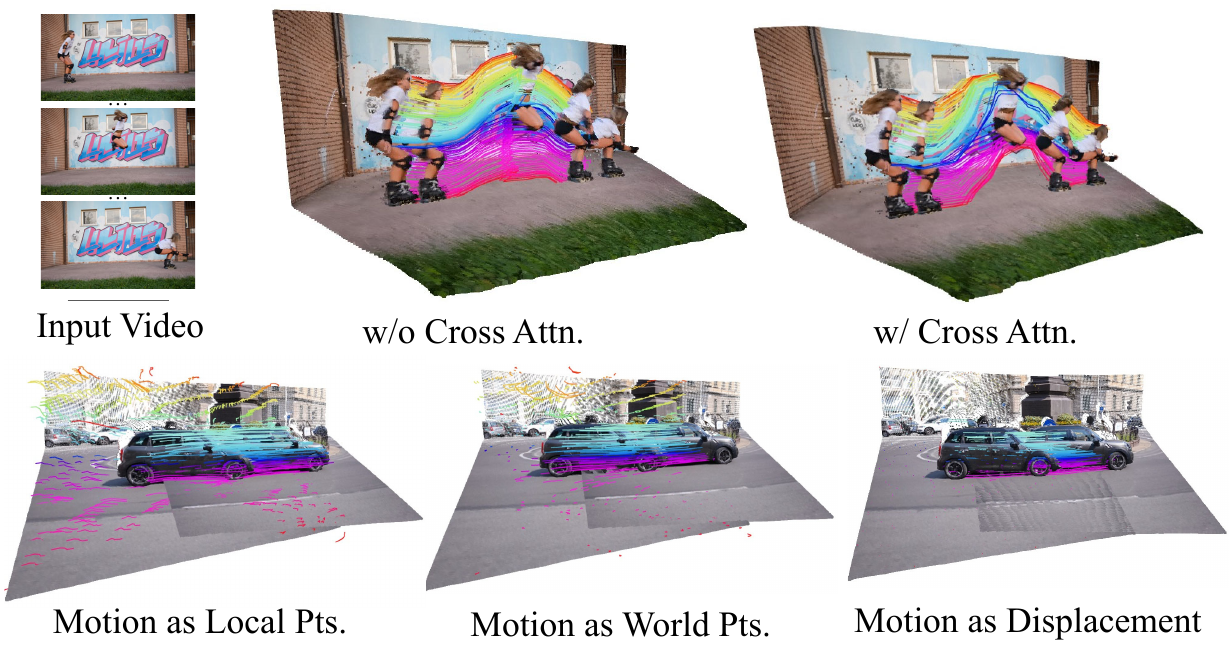}}
\caption{\textbf{Qualitative ablation visualizations.} The first row shows the effectiveness of cross-attention in the motion head: without it, although the model outputs rough trajectories, it fails to capture fine details such as the motion of the girl's legs and hands when she is at the peak of a jump. The second row illustrates that outputting motion as point clouds can lead to inconsistent trajectories as it requires re-predicting base geometry for each time step.}
\label{fig:ablation}
\end{center}
\vspace{-8.5mm}
\end{figure}

\textbf{Motion Head Design.}  
Our motion head enables motion querying from arbitrary input views at arbitrary target timestamps. 
To analyze the contribution of each component in the motion head, we construct several variants by removing individual modules: (i) cross-attention between query tokens and target-time latent features, (ii) self-attention, and (iii) time-token conditioned AdaLN. All variants use the same number of layers and have comparable parameter sizes. As shown in Table~\ref{tab:ablation} (a), removing any component consistently degrades performance, indicating that all modules are necessary for effective motion decoding. Among them, removing either attention module results in the largest performance drop. In Figure~\ref{fig:ablation}, we also quantitatively observe that without cross-attention, the decoder struggles to model complex non-rigid motions, such as hand and leg movements, producing over-smoothed trajectories that do not align with the true motion. This suggests that self-attention and adaptive normalization alone are insufficient for handling large and detailed temporal displacements, and direct access to target-time features is critical for accurate motion estimation.

\textbf{Factorized Motion.}  
We further evaluate the effectiveness of our factorized motion representation by comparing it with alternative output parameterizations commonly used in 3D reconstruction~\cite{wang2025vggt, wang2025pi}. Specifically, we replace our displacement-based formulation with two point-based variants: directly predicting 3D coordinates in (i) a shared world coordinate system, or (ii) each view's own camera coordinate system. As reported in Table~\ref{tab:ablation} (b), both point-based variants perform worse than our factorized representation. This performance gap arises mainly from differences in representation. Direct point prediction entangles geometry and motion in a single output space, forcing the network to jointly learn shape and temporal correspondences, which significantly increases learning difficulty. Qualitative results in Figure~\ref{fig:ablation} further support this observation. Our formulation explicitly decouples static geometry from time-dependent motion via displacement fields, reducing unnecessary recomputation of geometry and improving temporal consistency.

\vspace{-1mm}
\section{Conclusion}
\label{sec:conclusion}

We present \nickname, a unified feed-forward transformer framework for 4D reconstruction from monocular videos. Central to our approach is a novel \textit{encode-once, query-anywhere and anytime} paradigm, in which a compact 4D representation of the entire video is learned once and subsequently queried to recover geometry and motion at arbitrary time instances. This paradigm effectively bridges the global spatio-temporal modeling with flexible, on-demand query-based reconstruction, achieving both accurate 4D reconstruction and high efficiency. Extensive experiments demonstrate that \nickname consistently outperforms prior methods across a wide range of challenging 4D reconstruction benchmarks. Looking ahead, unified models such as \nickname, which jointly reason about geometry and motion, represent a promising direction toward more general-purpose perceptual systems.
\vspace{-10mm}

\clearpage

\section*{Impact Statement}
This paper presents work whose goal is to advance the field of machine learning, with a particular focus on 4D reconstruction. The proposed approach has the potential to benefit applications in robotics, augmented/virtual reality, and content creation. While the method may have many potential societal consequences, none of which we feel must be specifically highlighted here.

\bibliography{main.bib}
\bibliographystyle{icml2026}

\clearpage
\appendix
\onecolumn
\section*{Appendix}
\vspace{5mm}

\section{Additional Implementation Details}
\label{app:implementation}

\subsection{Architecture Details}
\label{sec:arch_details}

We adopt the ViT-Giant (ViT-G) architecture from DINOv2~\cite{oquab2023dinov2} as our encoder, which consists of 40 transformer layers with a feature dimension of 1,536 and employs 24 attention heads. The encoder weight is initialized from Depth Anything 3 (DA3)~\cite{depthanything3}. For the geometry head, we follow a dual-DPT~\cite{dpt,depthanything3} design equipped with a lightweight MLP as the camera head. For the motion head, we employ a transformer-based decoder consisting of 4 layers of alternating self- and cross-attention with a hidden dimension of 1,536 and 16 attention heads. To generate high-resolution dense motion outputs, we leverage a DPT~\cite{dpt} upsampling strategy where we extract the feature tokens from the 19-th, 27-th, 33-rd, and 39-th blocks of the encoder. We therefore apply the motion head to these layers, concatenate the resulting outputs, and fuse them through the DPT head to regress the final dense motion displacement field.

\subsection{Dataset Details}
\label{sec:dataset_details}

We train 4RC on 7 datasets covering both dynamic and static environments. Table~\ref{tab:datasets} details the statistics and sampling ratio of each dataset during training. For 3D motion learning, we leverage four dynamic datasets with ground-truth motion: PointOdyssey~\cite{zheng2023point}, Dynamic Replica~\cite{karaev2023dynamicstereo}, Waymo~\cite{waymo}, and Kubric~\cite{greff2021kubric}. The motion supervision in these datasets varies from dense motion to sparse trajectories. Specifically for Kubric, we curate two subsets: 4,000 clips from the MOVi-F release (24 frames each) with dense motion annotations, and 6,000 clips from the CoTracker3~\cite{cotracker3} rendered training set (120 frames each) with sparse trajectory annotations. To ensure high-quality geometric reconstruction on static backgrounds, we additionally include three static datasets: DL3DV~\cite{ling2024dl3dv}, ScanNet++~\cite{yeshwanth2023scannet++}, and MVS-Synth~\cite{DeepMVS}.

\begin{table}[h]
\centering
\caption{\textbf{Training dataset statistics.} We train 4RC on a mixture of 7 datasets. The motion annotation varies between dense maps and sparse trajectories depending on the dataset source. Static datasets naturally provide motion annotations, i.e., zero movement.}
\resizebox{\textwidth}{!}{%
\begin{tabular}{c|l|c|c|c|c|c}
\toprule
\textbf{Index} & \textbf{Dataset} & \textbf{Scene Type} & \textbf{Real / Synthetic} & \textbf{Dynamic / Static} & \textbf{Motion Annotation} & \textbf{Sampling (\%)} \\
\midrule
1 & PointOdyssey~\cite{zheng2023point} & Mixed & Synthetic & Dynamic & Sparse & 22.12 \\
2 & Dynamic Replica~\cite{karaev2023dynamicstereo} & Mixed & Synthetic & Dynamic & Sparse & 29.20 \\
3 & Waymo~\cite{waymo} & Outdoor & Real & Dynamic & Dense & 4.42 \\
4 & Kubric~\cite{greff2021kubric} & Object & Synthetic & Dynamic & Dense \& Sparse & 26.55 \\
5 & DL3DV~\cite{ling2024dl3dv} & Mixed & Real & Static & Dense & 8.85 \\
6 & ScanNet++~\cite{yeshwanth2023scannet++} & Indoor & Real & Static & Dense & 3.54 \\
7 & MVS-Synth~\cite{DeepMVS} & Outdoor & Synthetic & Static & Dense & 5.31 \\
\bottomrule
\end{tabular}%
}
\label{tab:datasets}
\vspace{-2mm}
\end{table}

\subsection{Training Details}

During training, we apply standard data augmentations, including Gaussian blur ($p=0.2$), ColorJitter ($p=0.1$), and RandomGrayscale ($p=0.05$). Video frames are sampled in strict temporal order with a random interval ranging from 1 to 5 frames.
For motion supervision, we adopt a probabilistic sampling strategy. Specifically, in 20\% of the training iterations, we supervise the model using all available motion ground truth. In the remaining 80\%, we employ sparse supervision by retaining only the top 20--30\% of points with the largest displacement magnitudes. Empirically, we find that this strategy filters out static or low-motion regions, prevents the dominance of zero-motion signals and accelerates convergence.
For the ray map loss $\mathcal{L}_{\text{ray}}$ and the camera parameter loss $\mathcal{L}_{\text{cam}}$ in Equation~\ref{eq:loss}, we adopt the loss formulation from DA3~\cite{depthanything3} for supervision.

\section{Additional Experiments and Results}
\subsection{Streaming Version of \nickname}
\label{sec:streaming_4rc}

\begin{figure*}[!t]
    \begin{center}
    \centerline{\includegraphics[width=\textwidth]{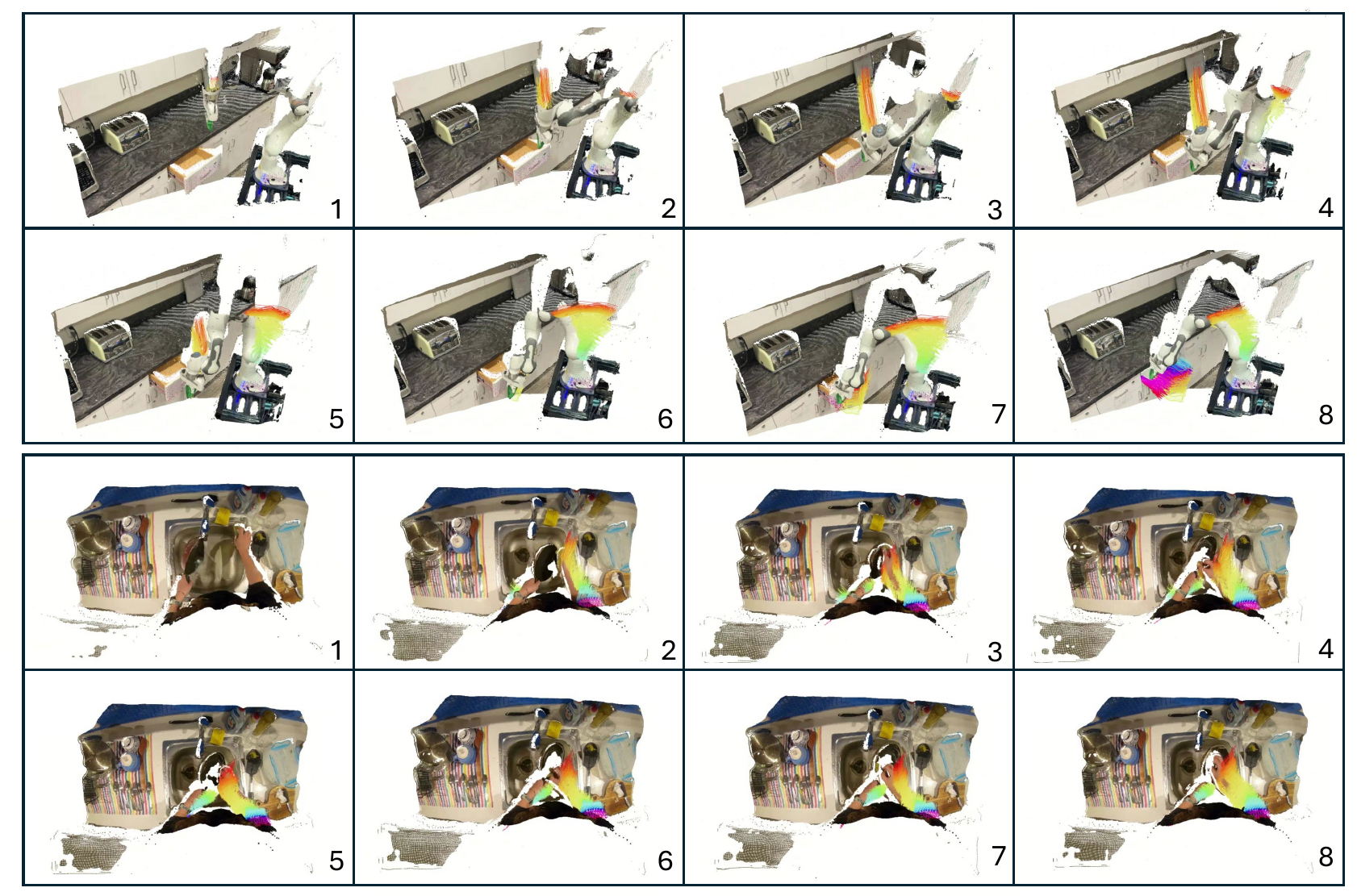}}
    \caption{\textbf{The visualization of S-\nickname results.} S-\nickname can infer 3D geometry and motion in an online manner, which is beneficial for downstream tasks such as robotic motion planning and egocentric understanding.}
    \label{fig:s4rc_vis}
    \end{center}
    \vspace{-4mm}
\end{figure*}

To support causal and online 4D reconstruction, we further introduce a streaming variant of \nickname (S-\nickname) which builds upon the STream3R~\cite{stream3r2025} architecture. Specifically, we replace our encoder with the pretrained STream3R backbone, which enforces unidirectional causal attention. The model is then fine-tuned using the proposed \nickname training objectives. Unlike standard \nickname, which processes the entire video in an offline manner, S-\nickname operates sequentially and achieves per-frame latency. We cache the 4D latent representation $\mathcal{F}$ for all processed frames. This enables flexible motion queries from the current view to any past timestamp, as well as point tracking from past views to the current time. As shown in Table~\ref{tab:s4rc_eval} and Figure~\ref{fig:s4rc_vis}, S-\nickname achieves competitive performance in 4D reconstruction while operating in an online manner, without access to global temporal context. Note that S-\nickname is trained for 20 epochs on 8 A100 GPUs.

\definecolor{best}{RGB}{255, 255, 255}
\definecolor{second}{RGB}{255, 255, 255}

\begin{table*}[h]
\centering
\caption{\textbf{4D reconstruction evaluation for S-4RC.}
S-\nickname enables online and streaming 4D reconstruction and achieves competitive performance compared to \nickname, even without access to global temporal context.
}

\label{tab:s4rc_eval}
\renewcommand{\arraystretch}{1.15}
\renewcommand{\tabcolsep}{1.2mm}
\resizebox{0.80\textwidth}{!}{%
\begin{tabular}{l cc cc cc cc cc cc}
\toprule
\multirow{3}{*}{Method} & \multicolumn{8}{c}{Point Tracking} & \multicolumn{4}{c}{Dense Tracking} \\
\cmidrule(lr){2-9} \cmidrule(lr){10-13}
  & \multicolumn{2}{c}{PO} & \multicolumn{2}{c}{DR} & \multicolumn{2}{c}{ADT} & \multicolumn{2}{c}{PStudio} & \multicolumn{2}{c}{Kubric} & \multicolumn{2}{c}{Waymo} \\
\cmidrule(lr){2-3} \cmidrule(lr){4-5} \cmidrule(lr){6-7} \cmidrule(lr){8-9} \cmidrule(lr){10-11} \cmidrule(lr){12-13} 
  & APD $\uparrow$ & EPE $\downarrow$ & APD $\uparrow$ & EPE $\downarrow$ & APD $\uparrow$ & EPE $\downarrow$ & APD $\uparrow$ & EPE $\downarrow$ & APD $\uparrow$ & EPE $\downarrow$ & APD $\uparrow$ & EPE $\downarrow$ \\
\midrule
S-\nickname & 73.29 & 0.3863 & 83.47 & 0.1970 & 86.12 & 0.1674 & 83.81 & 0.1795 & 75.60 & 1.168 & 46.02 & 1.971 \\
\nickname & \cellcolor{best}85.86 & \cellcolor{second}0.2498 & \cellcolor{best}88.65 & \cellcolor{best}0.1484 & \cellcolor{second}87.82 & \cellcolor{second}0.1480 & \cellcolor{second}87.32 & \cellcolor{second}0.1304 & \cellcolor{best}85.44 & \cellcolor{best}1.022 & \cellcolor{best}56.63 & \cellcolor{best}1.611 \\
\bottomrule
\end{tabular}
}
\end{table*}

\subsection{Additional Quantitative Evaluation on 4D Reconstruction}

As a complement to the evaluation in Table~\ref{tab:world_tracking_merged}, following WorldTrack~\cite{st4rtrack2025} and TAPVid-3D~\cite{koppula2024tapvid3d}, we apply global median scale alignment to match the predicted points with the ground truth. This alignment is feasible since both the predictions and the ground-truth points are represented in a shared world coordinate system defined by the camera of the first frame.
We additionally include a staged pipeline baseline composed of Monst3R~\cite{zhang2024monst3r} and SpaTracker~\cite{xiao2024spatialtracker}. 
Comprehensive evaluations in Table~\ref{tab:world_tracking_merged_median} demonstrate that our method outperforms approaches specifically designed for point tracking as well as concurrent 4D reconstruction methods, achieving state-of-the-art results on 4 out of 6 datasets.

\definecolor{best}{RGB}{180, 220, 200}
\definecolor{second}{RGB}{225, 250, 240}

\begin{table*}[h]
\centering
\caption{\textbf{4D reconstruction evaluation on tracking under global median scale alignment.}
As a complement to Table~\ref{tab:world_tracking_merged}, we further evaluate our method on dense-view tracking (a) and sparse-view tracking (b) under \textit{global median scale alignment} on dynamic datasets. Our method maintains strong performance across both evaluation protocols.
}

\label{tab:world_tracking_merged_median}
\renewcommand{\arraystretch}{1.15}
\renewcommand{\tabcolsep}{1.2mm}
\resizebox{0.95\textwidth}{!}{%
\begin{tabular}{l cc cc cc cc cc cc}
\toprule
\multirow{3}{*}{Method} & \multicolumn{4}{c}{(a) Dense Tracking} & \multicolumn{8}{c}{(b) Sparse Point Tracking} \\
\cmidrule(lr){2-5} \cmidrule(lr){6-13}
  & \multicolumn{2}{c}{Kubric} & \multicolumn{2}{c}{Waymo} & \multicolumn{2}{c}{PO} & \multicolumn{2}{c}{DR} & \multicolumn{2}{c}{ADT} & \multicolumn{2}{c}{PStudio} \\
\cmidrule(lr){2-3} \cmidrule(lr){4-5} \cmidrule(lr){6-7} \cmidrule(lr){8-9} \cmidrule(lr){10-11} \cmidrule(lr){12-13} 
  & APD $\uparrow$ & EPE $\downarrow$ & APD $\uparrow$ & EPE $\downarrow$ & APD $\uparrow$ & EPE $\downarrow$ & APD $\uparrow$ & EPE $\downarrow$ & APD $\uparrow$ & EPE $\downarrow$ & APD $\uparrow$ & EPE $\downarrow$ \\
\midrule
VGGT + CoTracker3~\cite{cotracker3} & - & - & - & - & 49.08 & 0.6532 & 74.73 & 0.2884 & 72.21 & 0.3548 & 66.28 & 0.3107 \\
Monst3R + SpaTracker~\cite{xiao2024spatialtracker} & - & - & - & - & 47.65 & 0.5917 &  55.49 & 0.8823 & 51.95 & 0.5362 & 50.16 & 0.4837 \\
SpaTrackerV2~\cite{xiao2025spatialtracker} & - & - & - & - & 69.57 & 0.3780 & 73.43 & 0.2732 & \cellcolor{best}92.22 & \cellcolor{best}0.0915 & 74.16 & 0.2272 \\
St4RTrack~\cite{st4rtrack2025}        & 35.33 & 3.465 & 2.51 & 10.139 & 67.95 & 0.3140 & 73.74 & 0.2682 & 76.01 & 0.2680 & 69.67 & 0.2637 \\
TraceAnything~\cite{liu2025traceanythingrepresentingvideo} & 27.37 & \cellcolor{second}1.952 & 2.06 & 12.564 & 39.83 & 1.0593 & 60.63 & 0.5758 & 75.65 & 0.2511 & 71.33 & 0.2727 \\
Any4D~\cite{karhade2025any4d}        & - & - & - & - & 60.86 & 0.4194 & 68.39 & 0.3012 & 56.71 & 0.4320 & 60.03 & 0.3344 \\
V-DPM~\cite{sucar2025vdpm}        & \cellcolor{second}52.22 & 3.131 & \cellcolor{second}31.67 & \cellcolor{second}1.957 & \cellcolor{second}79.79 & \cellcolor{best}0.1994 & \cellcolor{second}76.38 & \cellcolor{second}0.2378 & 66.06 & 0.3426 & \cellcolor{best}76.36 & \cellcolor{best}0.1957 \\
\midrule
\textbf{\nickname~(Ours)} & \cellcolor{best}55.38 & \cellcolor{best}1.525 & \cellcolor{best}39.55 & \cellcolor{best}1.864 & \cellcolor{best}80.27 & \cellcolor{second}0.2681 & \cellcolor{best}82.91 & \cellcolor{best}0.1889 & \cellcolor{second}84.28 & \cellcolor{second}0.1766 & \cellcolor{second}69.04 & \cellcolor{second}0.2603 \\
\bottomrule
\end{tabular}
}
\end{table*}

\subsection{Additional Quantitative Evaluation on Depth Estimation}

We additionally include the KITTI dataset~\cite{kitti} and extend the video depth evaluation in the main paper. We compare with a broader set of baselines, including single-frame depth methods Marigold~\cite{ke2024repurposing} and DepthAnything-V2~\cite{yang2024depthv2}, video depth methods NVDS~\cite{NVDS}, DepthCrafter~\cite{hu2024-DepthCrafter}, and ChronoDepth~\cite{shao2024learningtemporallyconsistentvideo}, and joint depth-and-pose estimation approaches Robust-CVD~\cite{barsan2018robust} and CausalSAM~\cite{zhang2022structure}. All results are aligned using per-sequence scale and shift, enabling a more comprehensive and fair comparison for video depth evaluation. As shown in Table~\ref{tab:depth_estimation_scale_shift}, our method significantly outperforms existing depth estimation approaches and achieves competitive performance compared to the dynamic 3D reconstruction method Pi3~\cite{wang2025pi}. Notably, our method is not trained on large-scale 3D reconstruction datasets and is able to model dynamic object motion, rather than focusing solely on 3D reconstruction.

\definecolor{best}{RGB}{180, 220, 200}
\definecolor{second}{RGB}{225, 250, 240}

\begin{table*}[h]
\centering
\caption{\textbf{Depth estimation on Bonn, Sintel, and KITTI datasets.} We compare a series of methods that explicitly predict video depth using per-sequence scale \& shift alignment.
}
\label{tab:depth_estimation_scale_shift}
\renewcommand{\arraystretch}{1.1}
\renewcommand{\tabcolsep}{0.8mm}
\resizebox{0.65\textwidth}{!}{
\begin{tabular}{lcc cc cc}
\toprule
\multirow{2}{*}{Method} & \multicolumn{2}{c}{Bonn} & \multicolumn{2}{c}{Sintel} & \multicolumn{2}{c}{KITTI} \\
\cmidrule(lr){2-3} \cmidrule(lr){4-5} \cmidrule(lr){6-7}
 & Rel $\downarrow$ & $\delta < 1.25 \uparrow$ & Rel $\downarrow$ & $\delta < 1.25 \uparrow$ & Rel $\downarrow$ & $\delta < 1.25 \uparrow$ \\
\midrule
Marigold~\cite{ke2024repurposing} & 0.091 & 93.1 & 0.532 & 51.5 & 0.149 & 79.6 \\
Depth-Anything-V2~\cite{yang2024depthv2} & 0.106 & 92.1 & 0.367 & 55.4 & 0.140 & 80.4 \\
NVDS~\cite{NVDS} & 0.167 & 76.6 & 0.408 & 48.3 & 0.253 & 58.8 \\
ChronoDepth~\cite{shao2024learningtemporallyconsistentvideo} & 0.100 & 91.1 & 0.687 & 48.6 & 0.167 & 75.9 \\
DepthCrafter~\cite{hu2024-DepthCrafter} & 0.075 & 97.1 & 0.292 & 69.7 & 0.110 & 88.1 \\
Robust-CVD~\cite{kopf2021rcvd} & - & - & 0.703 & 47.8 & - & - \\
CasualSAM~\cite{zhang2022structure} & 0.169 & 73.7 & 0.387 & 54.7 & 0.246 & 62.2 \\
DUSt3R-GA~\cite{wang2024dust3r} & 0.156 & 83.1 & 0.531 & 51.2 & 0.135 & 81.8 \\
MASt3R-GA~\cite{mast3r_arxiv24} & 0.167 & 78.5 & 0.327 & 59.4 & 0.137 & 83.6 \\
MonST3R-GA~\cite{zhang2024monst3r} & 0.066 & 96.4 & 0.333 & 59.0 & 0.157 & 73.8 \\
Spann3R~\cite{wang2024spann3r} & 0.157 & 82.1 & 0.508 & 50.8 & 0.207 & 73.0 \\
CUT3R~\cite{wang2025cut3r} & 0.074 & 94.5 & 0.540 & 55.7 & 0.106 & 88.7 \\
VGGT~\cite{wang2025vggt} & 0.049 & 97.2 & \cellcolor{best}0.202 & \cellcolor{second}72.7 & \cellcolor{second}0.057 & \cellcolor{second}96.6 \\
Pi3~\cite{wang2025pi} & \cellcolor{best}0.044 & \cellcolor{best}97.5 & \cellcolor{second}0.229 & \cellcolor{best}73.2 & \cellcolor{best}0.038 & \cellcolor{best}98.4 \\
\midrule
\textbf{\nickname~(Ours)} & \cellcolor{second}0.048 & \cellcolor{second}97.3 & 0.249 & 67.0 & 0.058 & 95.5 \\
\bottomrule
\end{tabular}
}
\end{table*}


\subsection{More Visualizations}
\label{fig:in_the_wild}
\begin{figure*}[!t]
    \begin{center}
    \centerline{\includegraphics[width=\textwidth]{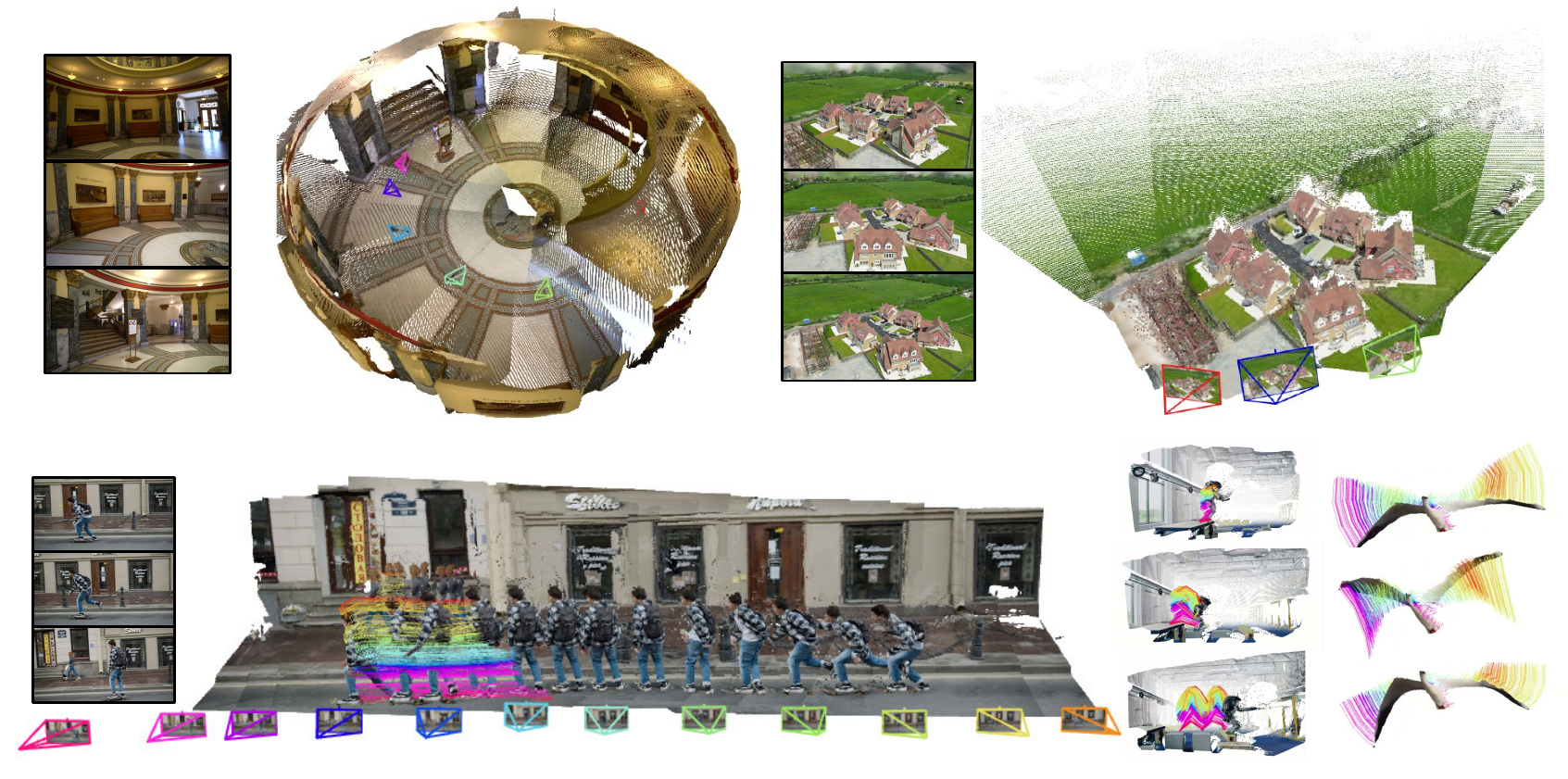}}
    \caption{\textbf{Visualization using \nickname on in-the-wild videos} of camera poses, static reconstruction, dynamic reconstruction, and 3D tracking.}
    \label{fig:supp_itw}
    \end{center}
    \vspace{-4mm}
\end{figure*}

We further provide additional visualizations of our \nickname results, including camera poses, static reconstruction, dynamic reconstruction, and 3D tracking on in-the-wild videos in Figure~\ref{fig:supp_itw}.

\subsection{Video Demo}
\label{subsec:demo_video}
We also provide a demo video on our  
\href{https://yihangluo.com/projects/4RC/}{project page}
to showcase the qualitative 4D reconstruction results of \nickname and S-\nickname.

\section{Limitations}

While our method achieves unified and flexible feed-forward 4D reconstruction and shows stronger performance than concurrent 4D reconstruction methods, several limitations remain. First, our approach struggles in scenarios where geometric recovery is inherently difficult. These include regions with extreme depth (e.g., distant clouds), transparent objects, or floating artifacts where the base geometry lacks sharp depth boundaries. We expect that improved depth estimation methods~\cite{xu2025pixel} and future advances in 3D reconstruction will help alleviate these issues. Second, we observe performance degradation in scenes with extreme or highly chaotic motion. This limitation mainly arises from the diversity of motion annotation in existing datasets, which provide insufficient supervision for such complex dynamics. Future work will explore scaling up training data to cover a broader range of motion patterns and kinematic diversity.

\end{document}